\documentclass[review]{elsarticle}
\usepackage{natbib}
\usepackage{lipsum}
\makeatletter
\def\ps@pprintTitle{%
 \let\@oddhead\@empty
 \let\@evenhead\@empty
 \def\@oddfoot{}%
 \let\@evenfoot\@oddfoot}
\makeatother
\usepackage{lscape}
\usepackage{graphicx}
\usepackage[font=small,skip=4pt]{caption}
\usepackage[table]{xcolor}
\usepackage{amsmath,amsfonts,amssymb}
\usepackage{blkarray}
\usepackage{algpseudocode}
\usepackage{tikz}
\usepackage{hyperref}
\usepackage{subcaption}
\usepackage{xcolor}
\usepackage[normalem]{ulem}
\usepackage{multirow}
\usepackage{array}

\newcolumntype{L}[1]{>{\raggedright\let\newline\\\arraybackslash\hspace{0pt}}m{#1}}
\newcolumntype{C}[1]{>{\centering\let\newline\\\arraybackslash\hspace{0pt}}m{#1}}
\newcolumntype{R}[1]{>{\raggedleft\let\newline\\\arraybackslash\hspace{0pt}}m{#1}}

\usepackage{mathtools}
\usepackage{algorithm}
\usepackage{algorithmicx}
\makeatletter
\def\BState{\State\hskip-\ALG@thistlm}
\makeatother

\newcommand{\comment}[1]{}
\usepackage{geometry}
\biboptions{authoryear}

\begin{document}

\begin{frontmatter}

\title{Forecasting directional movements of stock prices for intraday trading\\ using LSTM and random forests
	\footnote{
		\noindent
		\textbf{Acknowledgment:} The first author gratefully acknowledges the NTU-India Connect Research Internship Programme which allowed him to carry out part of this research project while visiting the Nanyang Technological University, Singapore.\\
		The second author gratefully acknowledges financial support by his Nanyang Assistant Professorship Grant (NAP Grant) \emph{Machine Learning based Algorithms in Finance and Insurance}.}
	}

\author[add1]{Pushpendu Ghosh}
\ead{f20150366@goa.bits-pilani.ac.in}
\author[add2]{Ariel Neufeld}
\ead{ariel.neufeld@ntu.edu.sg}
\author[add3]{Jajati Keshari Sahoo}
\ead{jksahoo@goa.bits-pilani.ac.in}

\address[add1]{Department of Computer Science $\&$ Information Systems, BITS Pilani K.K.Birla Goa campus, India}
\address[add2]{Division of Mathematical Sciences, Nanyang Technological University, Singapore}
\address[add3]{Department of Mathematics, BITS Pilani K.K.Birla Goa campus, India}

\begin{abstract}
	We employ both random forests  and LSTM networks (more precisely CuDNNLSTM)  as training methodologies to analyze their effectiveness in forecasting out-of-sample directional movements of constituent stocks of the S\&P 500 from January 1993 till December 2018 for intraday trading.
We introduce a multi-feature setting consisting not only of the returns with respect to the closing prices, but also with respect to the opening prices and intraday returns. 
As trading strategy, we use  \cite{krauss17} and \cite{krauss18} as benchmark. On each trading day, we buy the $10$ stocks with the highest probability and sell short the 10 stocks with the lowest probability 
to outperform  the market in terms of intraday returns 
-- all with equal monetary weight. 
Our empirical results show that the multi-feature setting provides a daily return, prior to transaction costs, of 0.64\% using LSTM networks, and 0.54\% using random forests. Hence we outperform the single-feature setting in \cite{krauss18} and \cite{krauss17}
consisting only of 
the daily returns with respect to the closing prices, having corresponding daily returns of  $0.41\%$ and of $0.39\%$ with respect to LSTM and random forests, respectively.\footnote{\cite{krauss18} and  \cite{krauss17} obtain $0.46\%$ and $0.43\%$, as the period from November 2015 until December 2018 was not included in their backtesting.}

\end{abstract}

\begin{keyword}
Random forest \sep LSTM  \sep Forecasting \sep Statistical Arbitrage \sep Machine learning \sep Intraday trading


\end{keyword}
\end{frontmatter}

\section{Introduction} 


In the last decade, machine learning methods have exhibited distinguished development in financial time series prediction. 
\cite{huck2009pairs} and \cite{huck2010pairs} construct statistical arbitrage strategies using Elman neural networks and a multi-criteria-decision method.
\cite{takeuchi2013} evolve a momentum trading strategy.
%
%
\cite{papaioannou2017s} develop a trend following trading strategy to forecast and trade S\&P 500  futures contracts.
\cite{tran2018}, \cite{sezer2018}, and \cite{singh2020feature} use neural networks for predicting time series data, whereas
\cite{borovykh2018} and \cite{xue2018} employ convolutional neural networks. 
Moreover, \cite{mallqui2019predicting} employ artificial neural networks, support vector machines, and ensemble algorithms to predict the direction as well as the maximum, minimum and closing prices of Bitcoin.
We also refer to \cite{harikrishnan2021machine} for a survey paper regarding the application of various machine learning algorithms for the prediction of stock prices.
%


In this paper, we focus on the application of long short-term memory networks (LSTM) and random forests to forecast directional movements of stock prices.
\cite{siami2018} compare LSTM with an autoregressive integrated moving average (ARIMA) model.
Their empirical results applied to financial data show that LSTM outperforms ARIMA in terms of lower forecast errors and higher accuracy.
The empirical results in \cite{qiu2020forecasting} using  LSTM demonstrate an improvement for stock price predictions when an attention mechanism is employed.
\cite{sharma2021use} observe that for both LSTM and autoregressive integrated moving average with exogenous variables (ARIMAX) models a considerable improvement of the prediction of stock price movements can be achieved when including a sentiment analysis.
\cite{moritz2014deep} employ random forests to predict returns of 
stocks using US CRSP data. As trading strategy, the top decile is bought, whereas the bottom one is sold short. 
\cite{lohrmann2019classification} construct a classification model using random forest to predict open-to-close returns of stocks in the S\&P 500.
\cite{basak2019predicting} use random forests and gradient boosted decision trees  (XGBoost), together with a selection of technical indicators, to analyze the performance 
for medium to long-run prediction of stock price returns,
and
\cite{sadorsky2021random} employs random forests to forecast the directions of stock prices of clean energy exchange traded funds.

Moreover,
\cite{krauss17}  compare different deep learning methods such as deep neural networks, gradient-boosted-trees and random forests. In a single-feature setting, the daily returns with respect to the closing prices of the S\&P~500
 from December~1992 until October~2015 
are provided to forecast one-day-ahead for every stock the probability of outperforming the market. As trading strategy, the $10$ stocks with the highest probability are bought and the 10 stocks with the lowest probability are sold short -- all with equal monetary weight.  It turns out that random forests achieve the highest return of each of the above deep learning methods with returns of 0.43\% per day, prior to transaction costs. \cite{krauss18} continue the study of \cite{krauss17} by employing LSTM networks as deep-learning  methodology and obtain returns of 0.46\% per day prior to transaction costs, therefore outperforming all the memory-free methods in \cite{krauss17}.

In our work, we use the results in \cite{krauss17} and \cite{krauss18} as benchmark for the ease of comparison. We introduce a multi-feature setting consisting not only of the returns with respect to the closing prices, but also with respect to the opening prices and intraday returns to predict for each stock, at the beginning of each day, the probability to outperform  the market in terms of intraday returns. As data set we use all stocks of the S$\&$P 500 from the period of January 1990 until December 2018.
We employ both random forests on the one hand and LSTM networks (more precisely CuDNNLSTM) on the other hand as training methodology and apply the same trading strategy as in  \cite{krauss17} and \cite{krauss18}. Our empirical results show that the multi-feature setting provides a daily return, prior to transaction costs, of 0.64\% for the LSTM network, and 0.54\% for the random forest, hence outperforming the single-feature setting in \cite{krauss18} and \cite{krauss17}, having corresponding daily returns of  $0.41\%$ and of $0.39\%$, respectively.\footnote{\cite{krauss18} and  \cite{krauss17} obtain $0.46\%$ and $0.43\%$, as the period from November 2015 until December 2018 was not included in their backtesting.}

The  remainder of this paper is organized as follows. 
In Section~\ref{sec:data} we explain the data sample as well as the software and  hardware we use. In Section~\ref{sec:method} we discuss the methodology we employ. The empirical results are then presented
  in Section~\ref{sec:result}.


\section{Data and technology}\label{sec:data}

 We collected adjusted closing prices and opening prices
  of all constituent stocks of the S$\&$P 500 from the period of January~1990 until December~2018 using Bloomberg. For each day, stocks with zero 
  volume were  not considered for trading at this day.

All experiments were executed in a NVIDIA Tesla V100 with 30 GB memory. The codes 
and simulations were implemented using Python 3.6.5 with a dependency of 
TensorFlow~1.14.0 and scikit-learn~0.20.4.
Visualization and statistical values were produced and calculated using the financial toolbox of MATLAB R2016b.

\section{Methodology}\label{sec:method}
Our methodology is composed of five steps. In the first step, we divide our raw data into 
study periods, where each study period is divided into a training part (for in-sample trading), and a trading part (for out-of-sample predictions). 
In the second step, we introduce our features, whereas 
in the third step we set up our targets. 
In the forth step, we  define the setup of our two machine learning methods we employ, namely random forest and CuDNNLSTM. 
Finally, in the fifth step, we establish a trading strategy for the trading part.


\subsection{Dataset creation with non-overlapping testing period}

We follow the procedure of \cite{krauss17} \& \cite{krauss18} and divide the  dataset consisting of 29 years starting from January 1990 till December 2018, using a 4-year window and 1-year stride, 
where each study period is divided into a training period of approximately 756 days ($\approx$ 3 years) and a trading period of  approximately 252 days ($\approx$ 1 year). As a consequence, we obtain 26 study periods with non-overlapping trading part. 

%
%
%
%
\begin{figure}[h!]
	\centering
	
	\includegraphics[width=.6\textwidth]{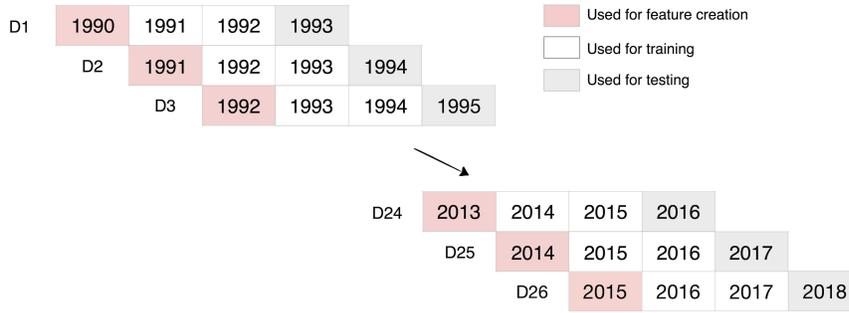} 
	\hfill
	\caption{Dataset creation with non-overlapping testing period}
	\label{fig:figure3.1.dataset}
\end{figure}

\subsection{Features selection}
\label{subsec:features}
%
Let $T_{study}$ denote the amount of days in a study period
and let $n_i$ be the number of stocks\footnote{We include stock prices of all S\&P500 constituents at the last day of the training data.} $s$ in $\textbf{\textit{S}}$ having complete historical data available at the end of each study period $i$. Moreover, we define the adjusted closing price and opening price of any stock $s \in \ $\textbf{\textit{S}} at time $t$  by 
$cp^{(s)}_{t}$
and
$op^{(s)}_{t}$ respectively.
\newline
Given a prediction day $t := \tau$, we have the following inputs and prediction tasks: \newline
\textit{Input}: We have the historical 
opening prices,  $op_t^{(s)},\ t\in \left \{ 0,1,...,\tau-1,\tau \right \}$, (including the prediction day's 
opening price $op_\tau^{(s)}$) 
as well as
 the historical adjusted closing prices $cp_t^{(s)},\ t\in \left \{ 0,1,...,\tau-1 \right \}$, (excluding the prediction day's closing price, $cp_\tau^{(s)}$).
 \\
\textit{Task}: Out of all $n$ stocks, predict $k$ stocks with the highest and $k$ stocks with the lowest intraday return $ir_{\tau,0} := \frac{cp_{\tau}}{op_{\tau}}-1$.
%
%

\newpage
\subsubsection{Feature  generation for Random Forest}\label{subsubsec:feature-RF}
For any stock $s \in \textbf{\textit{S}}$ and any time $t\in \{241,242,\dots,T_{study}\}$, the 
feature set we provide  to the random forest comprises of the following three signals:
\begin{enumerate}
    \item Intraday returns: $ir^{(s)}_{t,m} := \frac{cp^{(s)}_{t-m}}{op^{(s)}_{t-m}}-1$,
    \item Returns with respect to last closing price: $cr^{(s)}_{t,m} := \frac{cp^{(s)}_{t-1}}{cp^{(s)}_{t-1-m}}-1$,
    \item Returns with respect to opening price: $or^{(s)}_{t,m} := \frac{op^{(s)}_{t}}{cp^{(s)}_{t-m}}-1$,
\end{enumerate}
where $m \in \{1,2,3,...,20\} \cup \{40,60,80,...,240\}$, obtaining $93$ features, similar to \cite{takeuchi2013} \& \cite{krauss17}, who  considered the single feature case consisting of the simple return
$cr^{(s)}_{t,m} = \frac{cp^{(s)}_{t-1}}{cp^{(s)}_{t-1-m}}-1$. 
By the choice of $m\in\{1,2,3,...,20\}\cup \{40,60,80,...,240\}$,
 we consider in the first month the corresponding returns of each trading day, whereas for the subsequent 11 months we only consider the corresponding multi-period returns of each month.
No time series transformation such as, e.g., scaling or centering, is performed for the random forest.
\begin{figure}[h!]
	\centering
	
	\includegraphics[width=.6\textwidth]{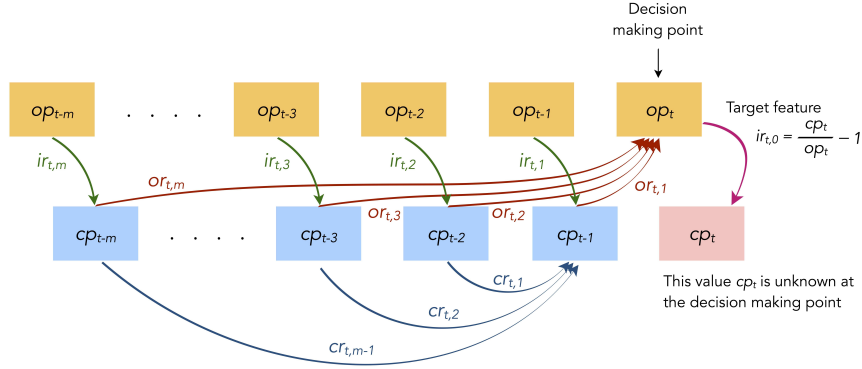} 
	\hfill
	\caption{Feature  generation for random forest}
	\label{fig:figure3.2.1.featureGenRandomForest}
\end{figure}
\subsubsection{Feature  generation for LSTM} 
Following the approach of \cite{krauss18}, but in a multi-feature setting rather than their single feature approach, we  
input the model with 240 timesteps and 3 features, and train it to predict the direction of the 241$^{st}$ intraday return. 

More precisely, for each stock $s$ at time $t$, we first consider the following three features, $ir^{(s)}_{t,1}$, $cr^{(s)}_{t,1}$, $or^{(s)}_{t,1}$ defined above in Subsection~\ref{subsubsec:feature-RF}. Then we apply the Robust Scaler standardization 
\begin{equation*}
\widetilde{f}^{(s)}_{t,1}:= \tfrac{{f}^{(s)}_{t,1} - \textup{Q}_2({f}^{(s)}_{\cdot,1})}{\textup{Q}_3({f}^{(s)}_{\cdot,1})-\textup{Q}_1({f}^{(s)}_{\cdot,1})},
\end{equation*}
where $\textup{Q}_1({f}^{(s)}_{\cdot,1})$, $\textup{Q}_2({f}^{(s)}_{\cdot,1})$ and $\textup{Q}_3({f}^{(s)}_{\cdot,1})$ are the first, second, and third quartile 
of ${f}^{(s)}_{\cdot,1}$, 
for each feature ${f}^{(s)}_{\cdot,1} \in \big\{ir^{(s)}_{\cdot,1}, cr^{(s)}_{\cdot,1},or^{(s)}_{\cdot,1}\big\}$ in the respective training period.
\\
 The Robust Scaler standardization (\cite{pedregosa2011scikit}) first subtracts (and hence removes) the median and then scales the data using the inter-quantile range, making it robust to  outliers.

Next, for each time $t\in \{241,242,\dots,T_{study}\}$, we generate overlapping sequences  of  240  consecutive, three-dimensional standardized features $\big\{\widetilde{F}^{(s)}_{t-239,1}, \widetilde{F}^{(s)}_{t-238,1}, \dots,$ $ \widetilde{F}^{(s)}_{t,1}\big\}$, where 
$\widetilde{F}^{(s)}_{t-i,1}:=\big(\widetilde ir^{(s)}_{t-i,1},\widetilde{cr}^{(s)}_{t-i,1},\widetilde{or}^{(s)}_{t-i,1}\big)$, $i\in \{239,238,\dots,0\}$.
\begin{figure}[H]
	\centering
	\includegraphics[width=.55\textwidth,trim={80 160 80 140},clip]{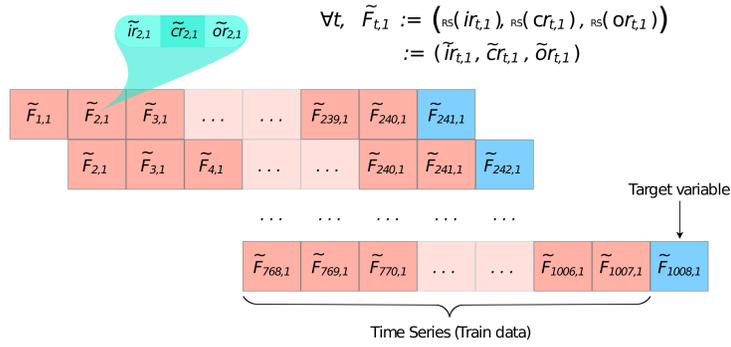}
	\hfill
	\caption{Feature  generation for LSTM}
	\label{fig:figure3.2.2.featureGenRandomLSTM}
\end{figure}

%
%
\subsubsection{Train-test split}

For each stock $s\in \textbf{\textit{S}}$, we create a matrix with $M$ columns and $T_{study}$ 
rows, where $M$ is the number of features. Hence $M$ is equal to 93 and $\left( 240, 3\right)$ when using random forest and LSTM, respectively. We fill the matrix with respective $M$ features as defined \hyperref[subsec:features]{above}. 
Since by definition, $ir_{t,m}$ is not defined when $t \leq m$, columns of the top 240 rows are partially filled and are hence removed. This removal leaves 
$T_{study}-240$ rows (i.e.\ for $t = \{241,242,243,...,T_{study}\}$) which is split into two parts, namely  approximately from $t=241$ to $t=756$, and  from $t=757$ to $t=T_{study}$, for training and testing purposes, respectively. Note that typically $T_{study}=1008$.
\par
At the end, we concatenate the training data of all stocks in $\textbf{\textit{S}}$ to get the collective training set. Hence the training set is a matrix with approximate size of $500 \times 516 = 258000$ rows (instances) and $M$ columns (features), along with their corresponding \hyperref[subsec:target]{target}, whereas the 
 trading set is a matrix with approximate size of $500 \times 252 = 126000$ rows (instances) and $M$ columns (features).
 \begin{figure}[h!]
 	\centering
 	
 	\includegraphics[width=.6\textwidth]{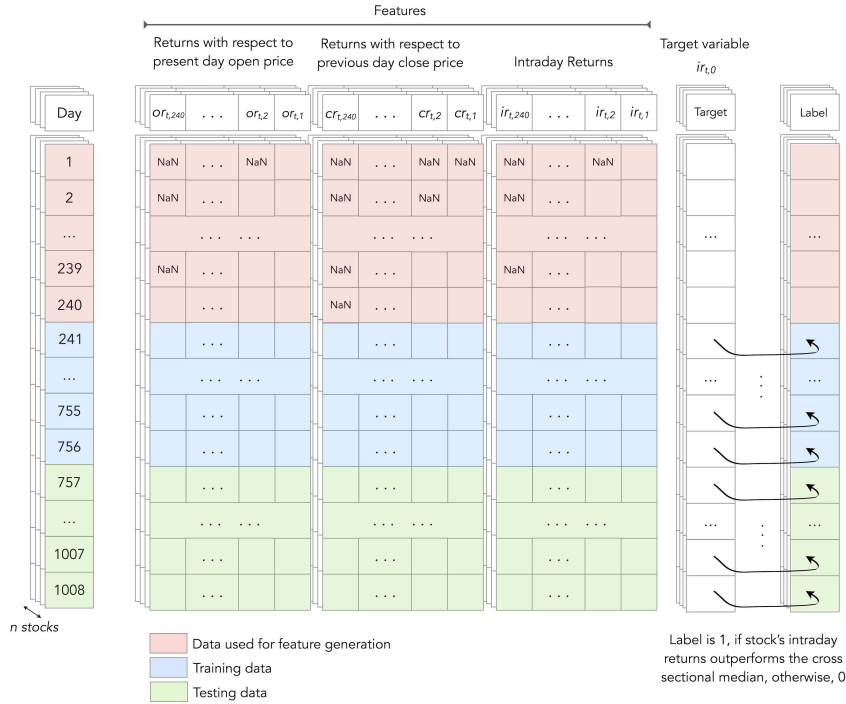} 
 	\hfill
 	\caption{Train-test split matrix}
 	\label{fig:figure3.2.3.train_est_split}
 \end{figure}
%
%
\subsection{Target selection}
\label{subsec:target}
Following \cite{takeuchi2013} and \cite{krauss18} we divide  each 
stock at time $t$  into 2 classes of equal size, based on their intraday returns $ir^{(s)}_{t,0}$. Class~0 is realized if $ir^{(s)}_{t,0}$ of stock $s$ is smaller than the cross-sectional median intraday return of all stocks at time $t$, whereas  Class~1 is realized if $ir^{(s)}_{t,0}$ of stock $s$ is bigger than the cross-sectional median intraday return of all stocks at time $t$.


\subsection{Model training specification}
\subsubsection{Model specification for Random forest}
As first model, we use random forests introduced by \cite{ho1995random} and expanded by \cite{breiman2001random},
with the following parameters: 
\begin{itemize}
	\item Number of decision trees in the forest = 1000 
	
	\item Maximum depth of each tree\footnote{Our empirical study from hyperparameter tuning suggests that forests with maximum depth of 10 give the highest accuracy.} = 10
	
	\item For every split, we select $m:= {\lfloor \sqrt{p} \rfloor}$ features randomly from the $p=93$ features in the data, see \cite{pedregosa2011scikit}. 
	
\end{itemize}
We refer to \cite[Subsection~4.3.3]{krauss17} and \cite[Subsection~3.4]{krauss18} for further details regarding random forests.
\subsubsection{Model specification for LSTM} 
LSTM is a recurrent neural network introduced by \cite{schmidhuber1997long}; we refer to \cite{krauss18} for a detailed description. 
Since  the training of LSTMs is very time consuming, and also to efficiently utilize the power of GPUs, we perform our experiments using CuDNNLSTMs (\cite{chetlur2014cudnn}). CUDA Deep Neural Network library (cuDNN) is a GPU-accelerated library for deep neural networks. We gain immense speedup (up to 7.2x, see \cite{braun2018lstm}) in training and predicting time. 
For the ease of comparison\footnote{We tested several values  
	for the amount of cells,
	as well as several LSTM architectures.
	They only marginally influence the empirical results.} 
we follow the network architecture used in \cite{krauss18}. We create a model with 25 cells 
of CuDNNLSTM, followed by a dropout layer of 0.1 and then a dense layer of 2~output nodes with softmax activation function. 

\begin{itemize}
    \item Loss function: categorical cross-entropy
    \item Optimizer: RMSProp (with the keras default learning rate of 0.001)
    \item Batch size: 512
    \item Early stopping: patience of 10 epochs, monitoring the validation loss
    \item Validation split: 0.2.
\end{itemize}
\subsection{Prediction and trading methodology}
We forecast the probability $\mathcal{P}^{(s)}_{t}$ for each stock $s$ to outperform the median intraday return $ir^{(s)}_{t,0}$. Next, as trading strategy, we follow \cite{krauss17} and \cite{krauss18}
and go long the top $k=10$ stocks with highest $\mathcal{P}^{(s)}_{t}$ and go short 
the worst $k=10$ stocks with lowest $\mathcal{P}^{(s)}_{t}$ -- all with equal monetary weight.  Each long and short transaction are subjected to 0.05$\%$ slippage cost on each half-turn, as suggested by \cite{avellaneda2010statistical}, so each day's transaction is penalized with a total of 0.2$\%$. 


\section{Results and Discussion}\label{sec:result}

The empirical results\footnote{All the codes are available on \url{https://github.com/pushpendughosh/Stock-market-forecasting}} 
show that our multi-feature setting consisting not only of the returns with respect to the closing prices, but also with respect to the opening prices and intraday returns, outperforms the single feature setting of \cite{krauss17} and \cite{krauss18}, both with respect to random forests and LSTM. We refer to  "IntraDay" for our setting and "NextDay" for the setting in \cite{krauss17} and \cite{krauss18} in Tables~\ref{table:time}--\ref{table:after_cost} and Figures~\ref{fig:figure1}--\ref{fig:figure3}.

Indeed, our setting 
involving LSTM obtains, prior to transaction costs, a daily return of 0.64\%, compared to the setting in \cite{krauss18} obtaining a 0.41\% daily return. Also for random forests, our setting obtains a higher daily return of 0.54\%, compared to 0.39\% when using the setting as in \cite{krauss17}. The share of positive returns is at 69.67\% and 65.85\%  for LSTM and random forests. In addition,
our setting 
obtains 
higher sharpe ratio and lower standard deviation (i.e.\ typical annualized risk-return metrics) in comparison with the one in \cite{krauss17} and \cite{krauss18}. 
Furthermore, 
our setting produces a lower maximum drawdown and lower daily value at risk (VaR); we refer to Tables~\ref{table:main}~\&~\ref{table:after_cost}. 

We also see that in our setting LSTM outperforms random forests, which is in line with the results of \cite{krauss18} showing that LSTM has an advantage compared to the  memory-free methods analyzed in \cite{krauss17}.

In Figures~\ref{fig:figure1}--\ref{fig:figure3}, we have divided the time period from January 1993 until December 2018 into three time-periods, analog to \cite{krauss18} and similar to \cite{krauss17}. Roughly speaking, the first time-period corresponds to a strong performance caused by, among others, the dot-com-bubble, followed by the time-period of moderation with the bursting of the dot-com bubble and the financial crisis of 2008, ending with the time-period of deterioration; probably since by that time on, machine learning algorithms are broadly available and hence diminishes the opportunity of creating statistical arbitrage having a technological advantage. We refer to \cite{krauss17} and \cite{krauss18} for a detailed discussion of these sub-periods. We see in Figures~\ref{fig:figure1}--\ref{fig:figure3} that in each of these sub-periods, our setting outperforms the one in \cite{krauss17} and \cite{krauss18}.

To show the importance of using three features instead of having a single feature, we additionally analyze in Tables~\ref{table:main}~\&~\ref{table:after_cost} 
the performance in the case of intraday-trading, but where only intraday returns $ir_{t,m}^{(s)}$ as a single feature is used.  
The experimental results show massive improvement in all metrics when using the three features introduced in Subsection~\ref{subsec:features}.
\\

As an outlook for future research, we remark  that the problem of forecasting directional movements of stocks (or currencies)  can be formulated as a reinforcement learning problem, both model-free and model-based. An interesting analysis would be to compare these different reinforcement learning methodologies to both LSTM and random forests, particularly applied to cryptocurrencies. We leave this open for future studies.
\vspace{0.9cm}
\begin{table}[H]
	\scriptsize
	\centering
	\renewcommand{\arraystretch}{1}
	\begin{tabular}{ p{4cm} || >{\centering\arraybackslash}p{1.2cm} >{\centering\arraybackslash}p{1.2cm} >{\centering\arraybackslash}p{1.2cm} >{\centering\arraybackslash}p{1.2cm}  >{\centering\arraybackslash}p{1.2cm} >{\centering\arraybackslash}p{1.2cm} }

		\rowcolor{lightgray}
		
		Metric & 3-Feature IntraDay LSTM & 3-Feature IntraDay RF & 1-Feature NextDay LSTM & 1-Feature NextDay RF & 1-Feature IntraDay LSTM & 1-Feature IntraDay RF \\
		Time per epoch (in sec) & 33.1  & - & 166  & - & 13.8 & - \\  
		Training time (in min) & 24.21  & 7.21  & 112.3  & 2.59 & 10.4 & 2.56 \\
		Decision making time (in sec) & 0.086924 & 0.419563 & 0.180778 & 0.380040 & 0.036128 & 0.374121 \\
		\hline
		
	\end{tabular}
	\caption{Time comparison}
	\label{table:time}
\end{table}
%
%
\begin{table}[H]
	\scriptsize
	\centering
	\renewcommand{\arraystretch}{1}
	\begin{tabular}{ p{2cm} || >{\centering\arraybackslash}p{1.2cm} >{\centering\arraybackslash}p{1.2cm} >{\centering\arraybackslash}p{1.2cm} >{\centering\arraybackslash}p{1.2cm} | >{\centering\arraybackslash}p{1.2cm} >{\centering\arraybackslash}p{1.2cm} | >{\centering\arraybackslash}p{1.2cm}}

		\rowcolor{lightgray}
		
		Metric $\phantom{for daily }$ of daily returns & 3-Feature IntraDay LSTM & 3-Feature IntraDay RF & 1-Feature NextDay LSTM & 1-Feature NextDay RF & 1-Feature IntraDay LSTM & 1-Feature IntraDay RF & SP500 Index \\
Mean (long) & 0.00332 & 0.00273 & 0.00257 & 0.00259 & 0.00094 & 0.00104 & 0.00033 \\
Mean (short) & 0.00312 & 0.00266 & 0.00158 & 0.00130 & 0.00180 & 0.00187 & 0.00000 \\
Mean return & 0.00644 & 0.00539 & 0.00414 & 0.00389 & 0.00274 & 0.00290 & 0.00033 \\
Standard error & 0.00019 & 0.00020 & 0.00024 & 0.00023 & 0.00021 & 0.00021 & 0.00014 \\
Minimum & -0.1464 & -0.1046 & -0.1713 & -0.1342 & -0.1565 & -0.1487 & -0.0903 \\
Quartile 1 & -0.0017 & -0.0028 & -0.0052 & -0.0051 & -0.0054 & -0.0050 & -0.0044 \\
Median & 0.00559 & 0.00462 & 0.00352 & 0.00287 & 0.00242 & 0.00221 & 0.00056 \\
Quartile 3 & 0.01433 & 0.01306 & 0.01294 & 0.01161 & 0.01086 & 0.01036 & 0.00560 \\
Maximum & 0.14101 & 0.14153 & 0.19884 & 0.28139 & 0.13896 & 0.16064 & 0.11580 \\
Share $>$ 0 & 0.69663 & 0.65857 & 0.60598 & 0.59479 & 0.58405 & 0.58937 & 0.53681 \\
Std. deviation & 0.01572 & 0.01597 & 0.01961 & 0.01831 & 0.01713 & 0.01683 & 0.01133 \\
Skewness & 0.15599 & 0.28900 & 0.36822 & 1.41199 & -0.1828 & 0.12051 & -0.1007 \\
Kurtosis & 9.71987 & 8.32627 & 10.8793 & 19.8349 & 10.1893 & 11.7758 & 11.9396 \\
\hline
1-percent VaR & -0.0352 & -0.0364 & -0.0492 & -0.0432 & -0.0461 & -0.0448 & -0.0313 \\
1-percent CVaR & -0.0519 & -0.0528 & -0.0712 & -0.0592 & -0.0678 & -0.0660 & -0.0451 \\
5-percent VaR & -0.0157 & -0.0170 & -0.0234 & -0.0208 & -0.0214 & -0.0197 & -0.0177 \\
5-percent CVaR & -0.0284 & -0.0297 & -0.0401 & -0.0345 & -0.0377 & -0.0357 & -0.0270 \\
Max. drawdown & 0.22345 & 0.19779 & 0.42551 & 0.23155 & 0.35645 & 0.43885 & 0.56775 \\
\hline
Avg return p.a. & 3.84750 & 2.75103 & 1.68883 & 1.53806 & 0.91483 & 1.00281 & 0.06975 \\
Std dev. p.a. & 0.24957 & 0.25358 & 0.31135 & 0.29071 & 0.27193 & 0.26719 & 0.17990 \\
Down dev. p.a. & 0.17144 & 0.17301 & 0.21204 & 0.18690 & 0.19270 & 0.18530 & 0.12970 \\
Sharpe ratio & 6.34253 & 5.20303 & 3.22732 & 3.23339 & 2.39560 & 2.59188 & 0.24867 \\
Sortino ratio & 62.7403 & 49.6764 & 27.8835 & 30.0753 & 19.0217 & 21.2964 & 1.77234 \\
\hline
		
	\end{tabular}
	\caption{Average performance metrics 
		of daily returns
		before transaction cost}
	\label{table:main}
\end{table}
%
\vspace{0.5cm}
\begin{table}[H]
	\scriptsize
	\centering
	\renewcommand{\arraystretch}{1}
	\begin{tabular}{ p{2cm} || >{\centering\arraybackslash}p{1.2cm} >{\centering\arraybackslash}p{1.2cm} >{\centering\arraybackslash}p{1.2cm} >{\centering\arraybackslash}p{1.2cm} | >{\centering\arraybackslash}p{1.2cm} >{\centering\arraybackslash}p{1.2cm} | >{\centering\arraybackslash}p{1.2cm}}

		\rowcolor{lightgray}
		
		Metric $\phantom{for daily }$ of daily returns & 3-Feature IntraDay LSTM & 3-Feature IntraDay RF & 1-Feature NextDay LSTM & 1-Feature NextDay RF & 1-Feature IntraDay LSTM & 1-Feature IntraDay RF & SP500 Index \\
Mean (long) & 0.00232 & 0.00173 & 0.00157 & 0.00159 & -0.0000 & 0.00004 & 0.00033 \\
Mean (short) & 0.00212 & 0.00166 & 0.00058 & 0.00030 & 0.00080 & 0.00087 & 0.00000 \\
Mean return & 0.00444 & 0.00339 & 0.00214 & 0.00189 & 0.00074 & 0.00090 & 0.00033 \\
Standard error & 0.00019 & 0.00020 & 0.00024 & 0.00023 & 0.00021 & 0.00021 & 0.00014 \\
Minimum & -0.1484 & -0.1066 & -0.1733 & -0.1362 & -0.1585 & -0.1507 & -0.0903 \\
Quartile 1 & -0.0037 & -0.0048 & -0.0072 & -0.0071 & -0.0074 & -0.0070 & -0.0044 \\
Median & 0.00359 & 0.00262 & 0.00152 & 0.00087 & 0.00042 & 0.00021 & 0.00056 \\
Quartile 3 & 0.01233 & 0.01106 & 0.01094 & 0.00961 & 0.00886 & 0.00836 & 0.00560 \\
Maximum & 0.13901 & 0.13953 & 0.19684 & 0.27939 & 0.13696 & 0.15864 & 0.11580 \\
Share $>$ 0 & 0.63129 & 0.59319 & 0.54279 & 0.53006 & 0.51534 & 0.50810 & 0.53681 \\
Std. deviation & 0.01572 & 0.01597 & 0.01961 & 0.01831 & 0.01713 & 0.01683 & 0.01133 \\
Skewness & 0.15599 & 0.28900 & 0.36822 & 1.41199 & -0.1828 & 0.12051 & -0.1007 \\
Kurtosis & 9.71987 & 8.32627 & 10.8793 & 19.8349 & 10.1893 & 11.7758 & 11.9396 \\
\hline
1-percent VaR & -0.0372 & -0.0384 & -0.0512 & -0.0452 & -0.0481 & -0.0468 & -0.0313 \\
1-percent CVaR & -0.0539 & -0.0548 & -0.0732 & -0.0612 & -0.0698 & -0.0680 & -0.0451 \\
5-percent VaR & -0.0177 & -0.0190 & -0.0254 & -0.0228 & -0.0234 & -0.0217 & -0.0177 \\
5-percent CVaR & -0.0304 & -0.0317 & -0.0421 & -0.0365 & -0.0397 & -0.0377 & -0.0270 \\
Max. drawdown & 0.39227 & 0.40139 & 0.87232 & 0.92312 & 0.97263 & 0.87123 & 0.56775 \\
\hline
Avg return p.a. & 1.94325 & 1.27179 & 0.63060 & 0.53901 & 0.16046 & 0.21146 & 0.06975 \\
Std dev. p.a. & 0.24957 & 0.25358 & 0.31135 & 0.29071 & 0.27193 & 0.26719 & 0.17990 \\
Down dev. p.a. & 0.17144 & 0.17301 & 0.21204 & 0.18690 & 0.19270 & 0.18530 & 0.12970 \\
Sharpe ratio & 4.32307 & 3.21553 & 1.60856 & 1.49969 & 0.54218 & 0.70557 & 0.24867 \\
Sortino ratio & 39.0873 & 27.9810 & 12.9274 & 12.7950 & 3.98288 & 5.34773 & 1.77234 \\
\hline

	\end{tabular}
	\caption{Average performance metrics
		of daily returns
		 after transaction cost}
	\label{table:after_cost}
\end{table}
%
%
%
\newpage
\begin{landscape}
\begin{figure}[p]
\centering

\includegraphics[width=.4\textwidth]{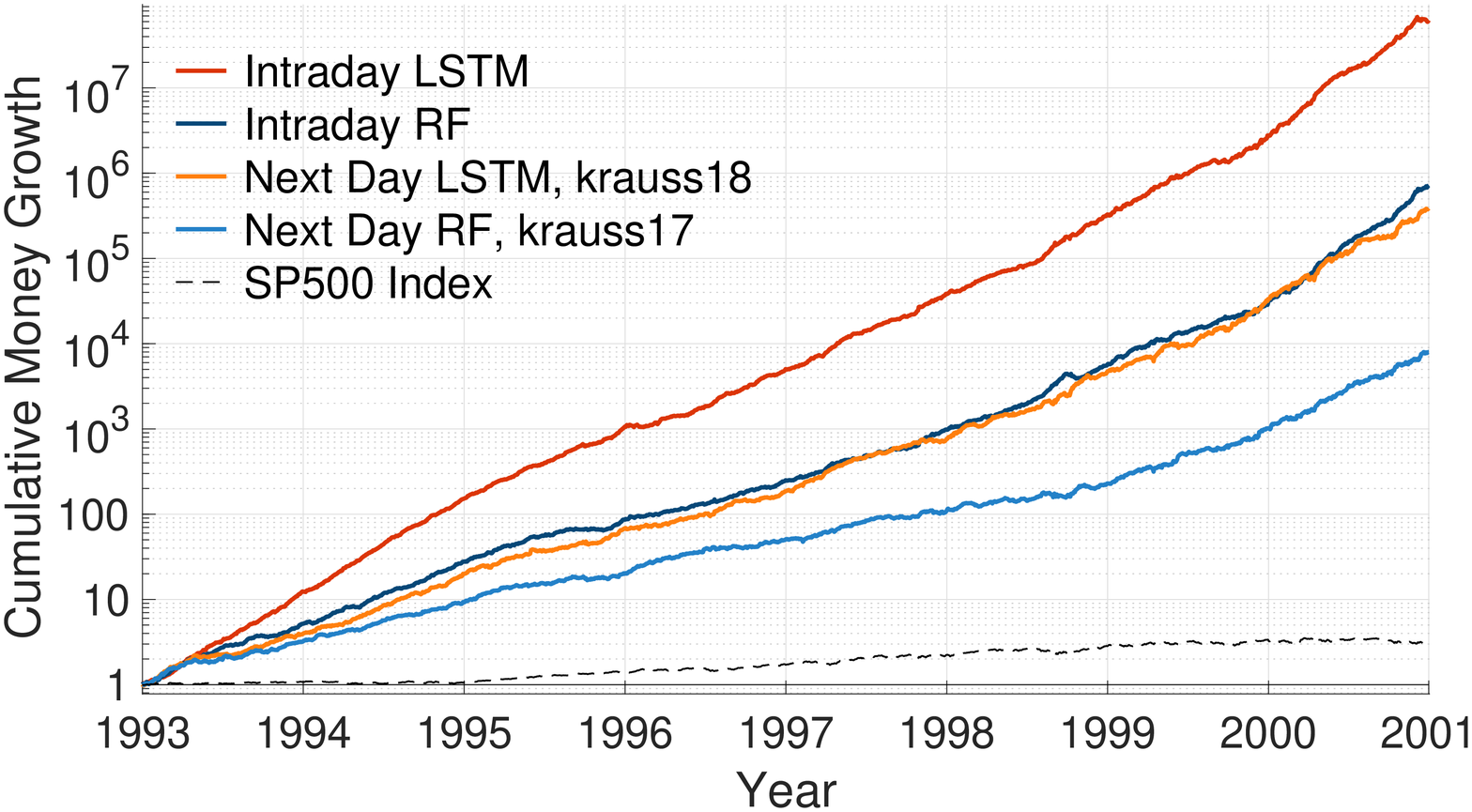} 
\hfill
\includegraphics[width=.4\textwidth]{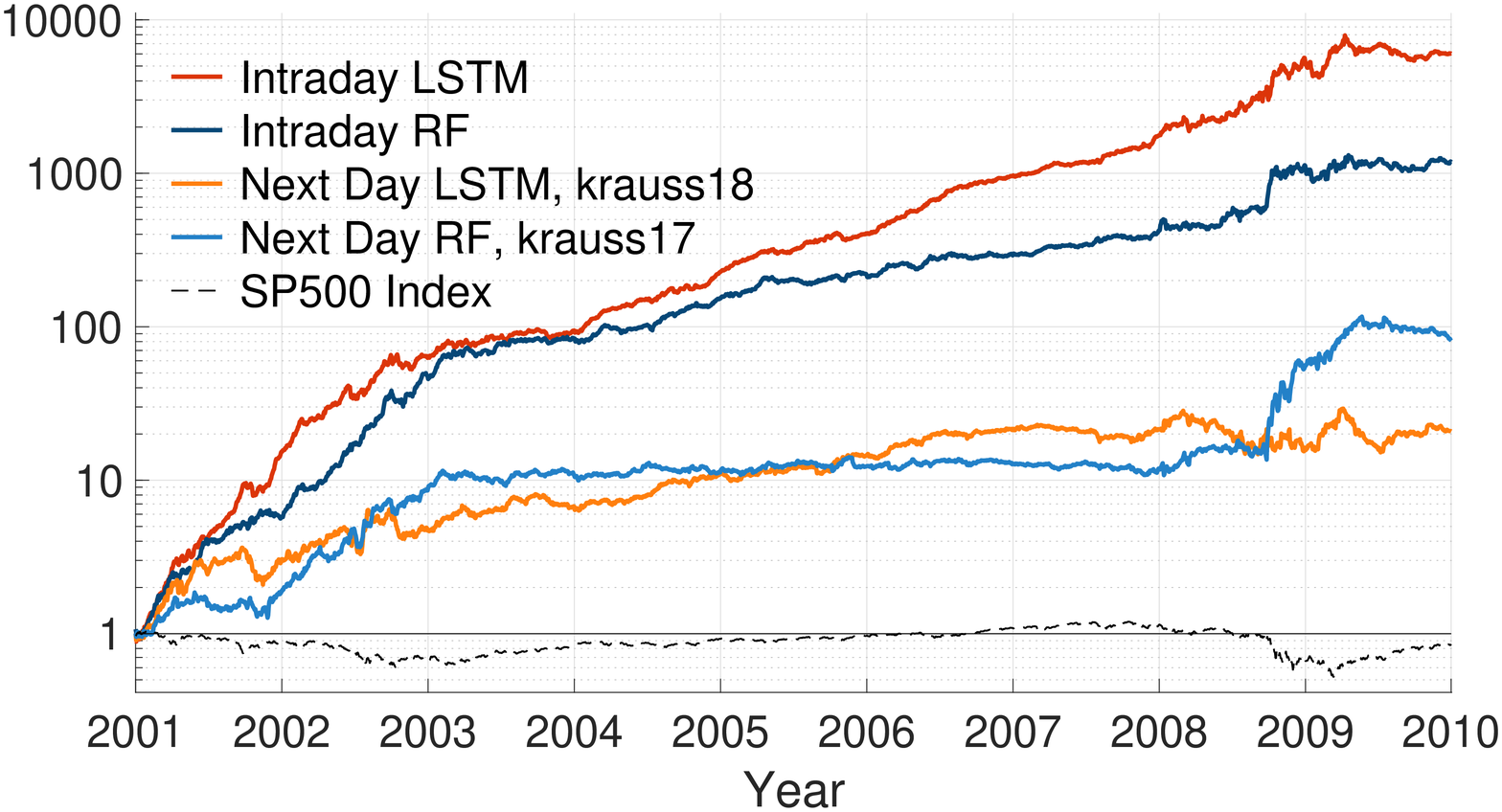} 
\hfill
\includegraphics[width=.4\textwidth]{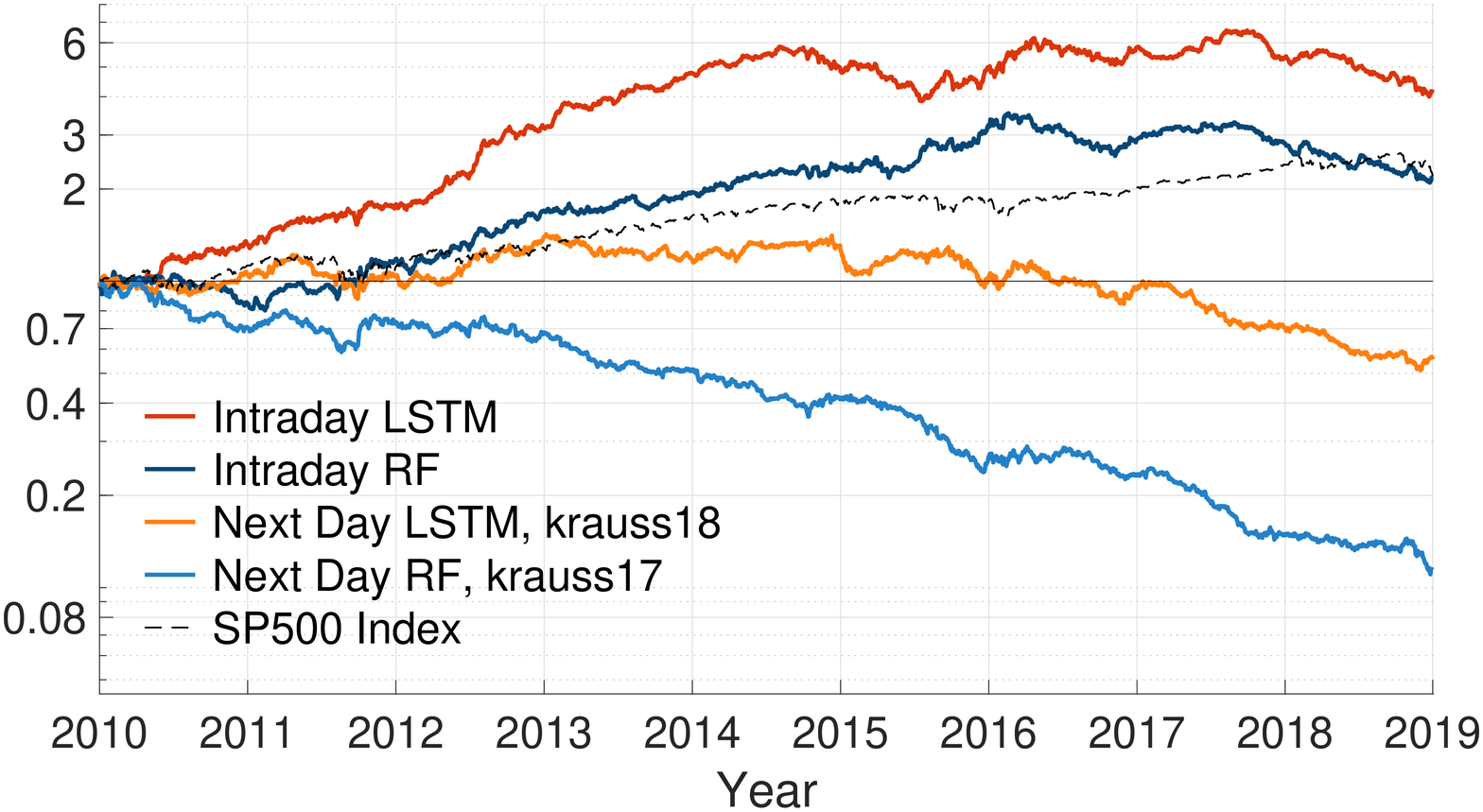} 
\caption{Cumulative money growth with US$\$$1 initial investment, after deducting transaction cost}
\label{fig:figure1}
\end{figure}

\begin{figure}[p]
\centering
\includegraphics[width=.4\textwidth]{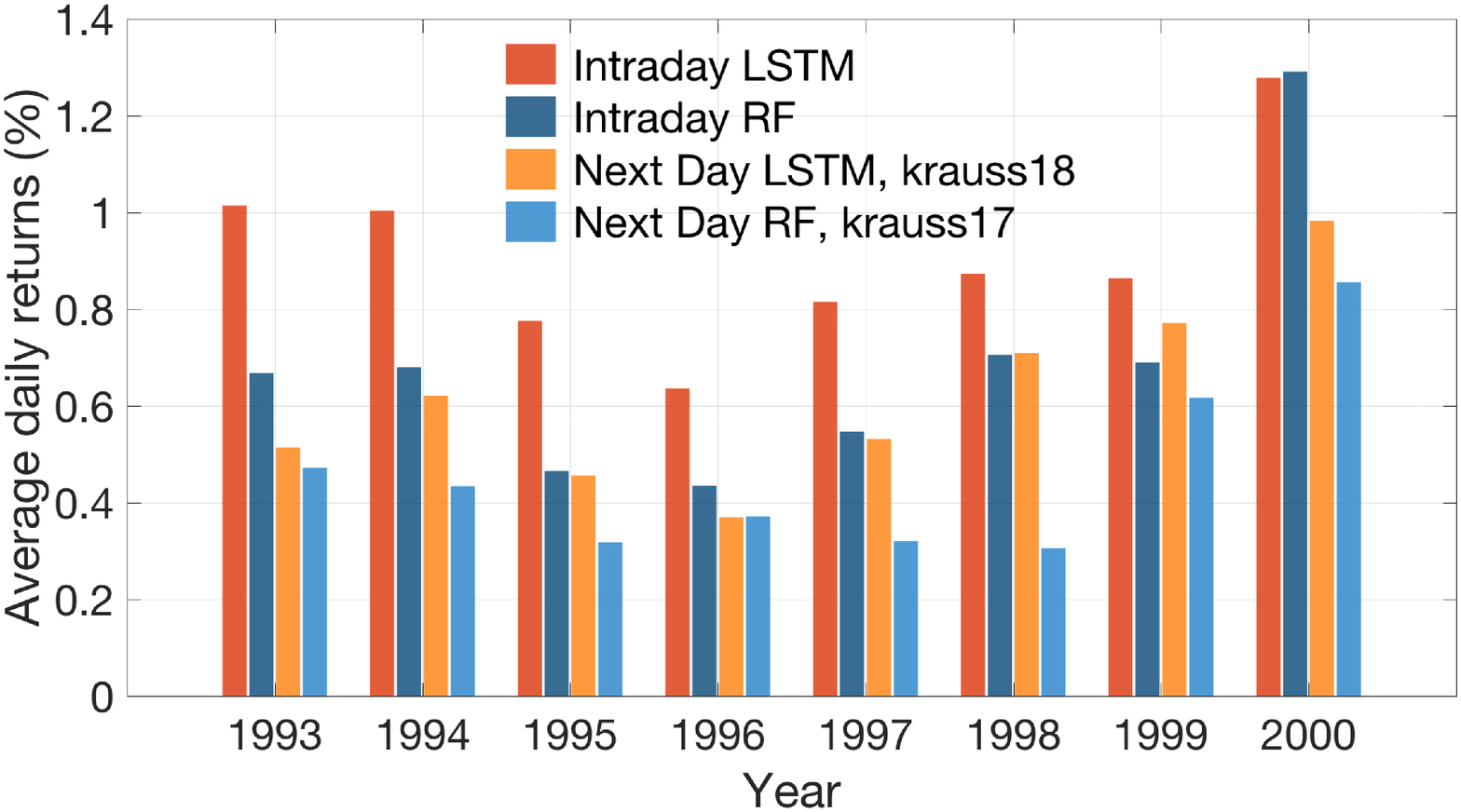} 
\hfill
\includegraphics[width=.4\textwidth]{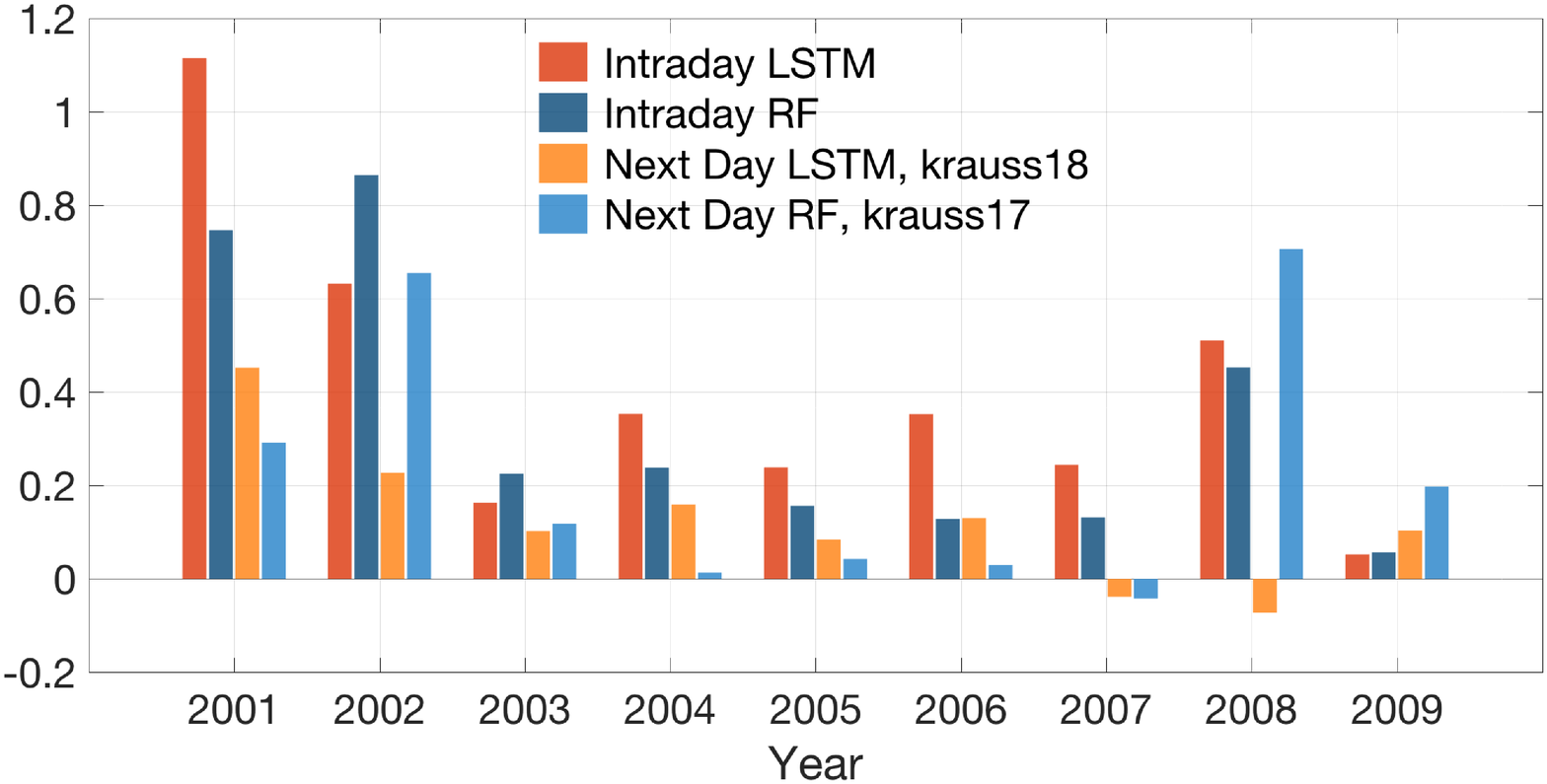} 
\hfill
\includegraphics[width=.4\textwidth]{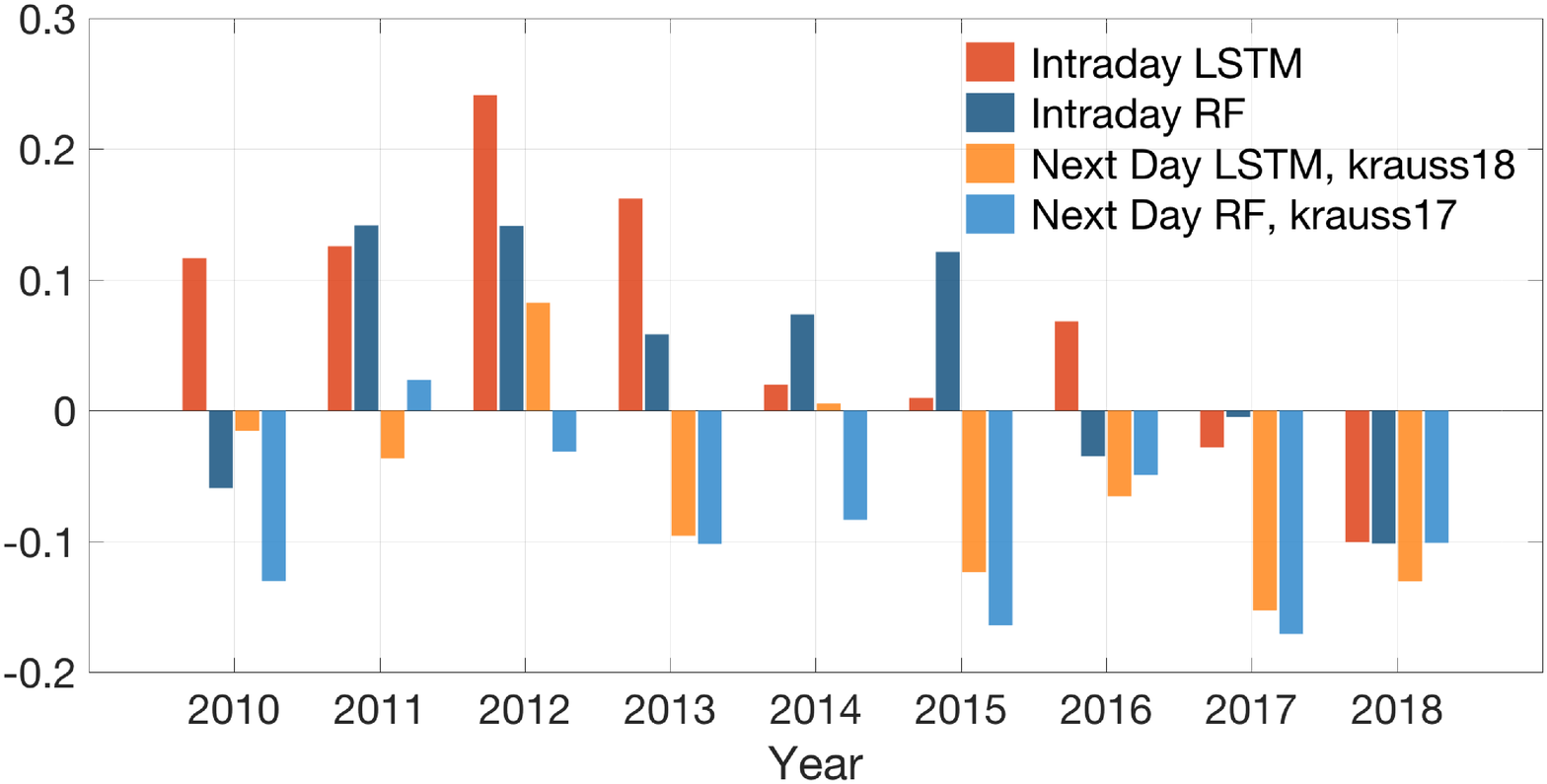} 
\caption{Average of daily mean returns, after deducting transaction cost}
\label{fig:figure2}
\end{figure}

\begin{figure}[p]
\centering
\includegraphics[width=.4\textwidth]{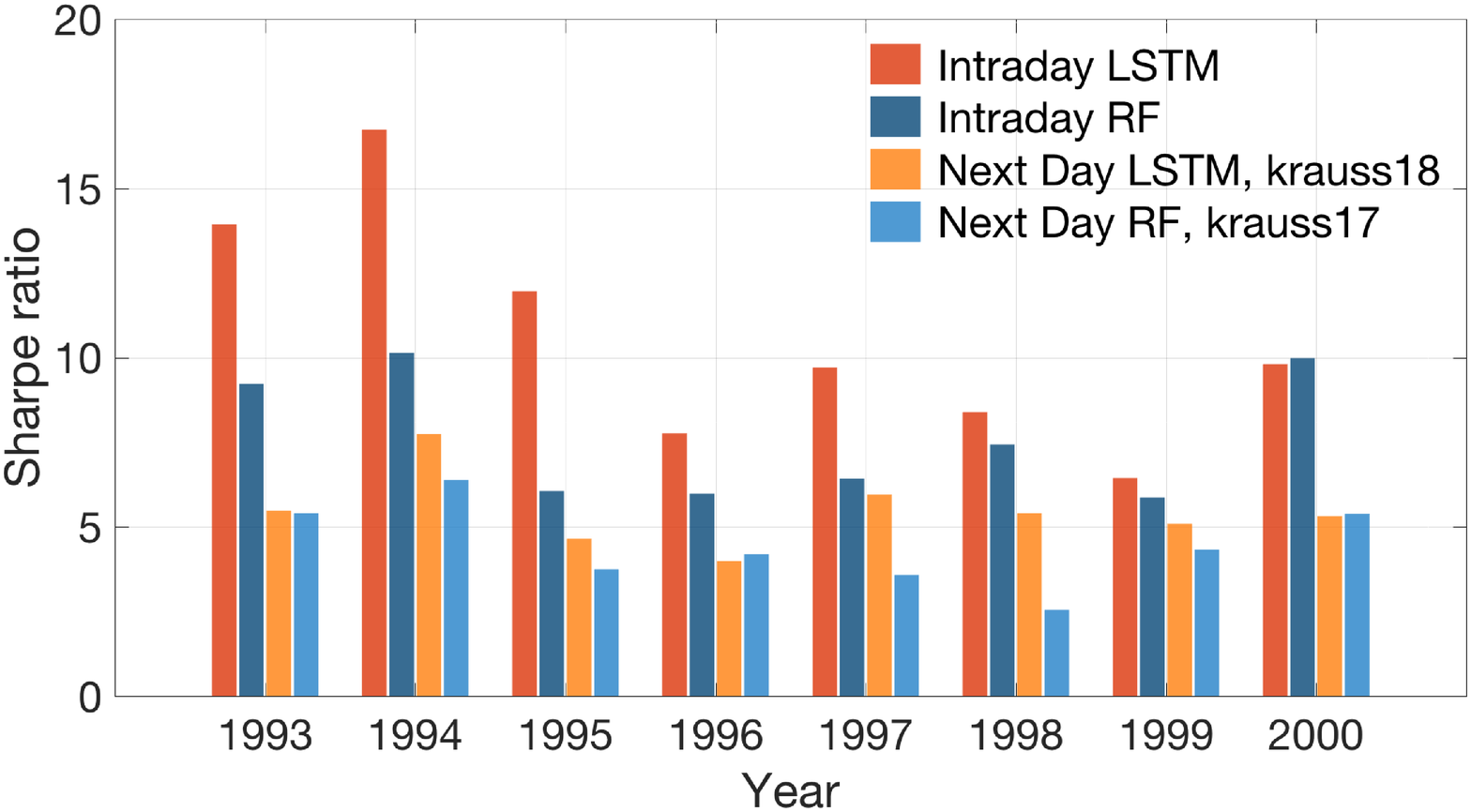} 
\hfill
\includegraphics[width=.4\textwidth]{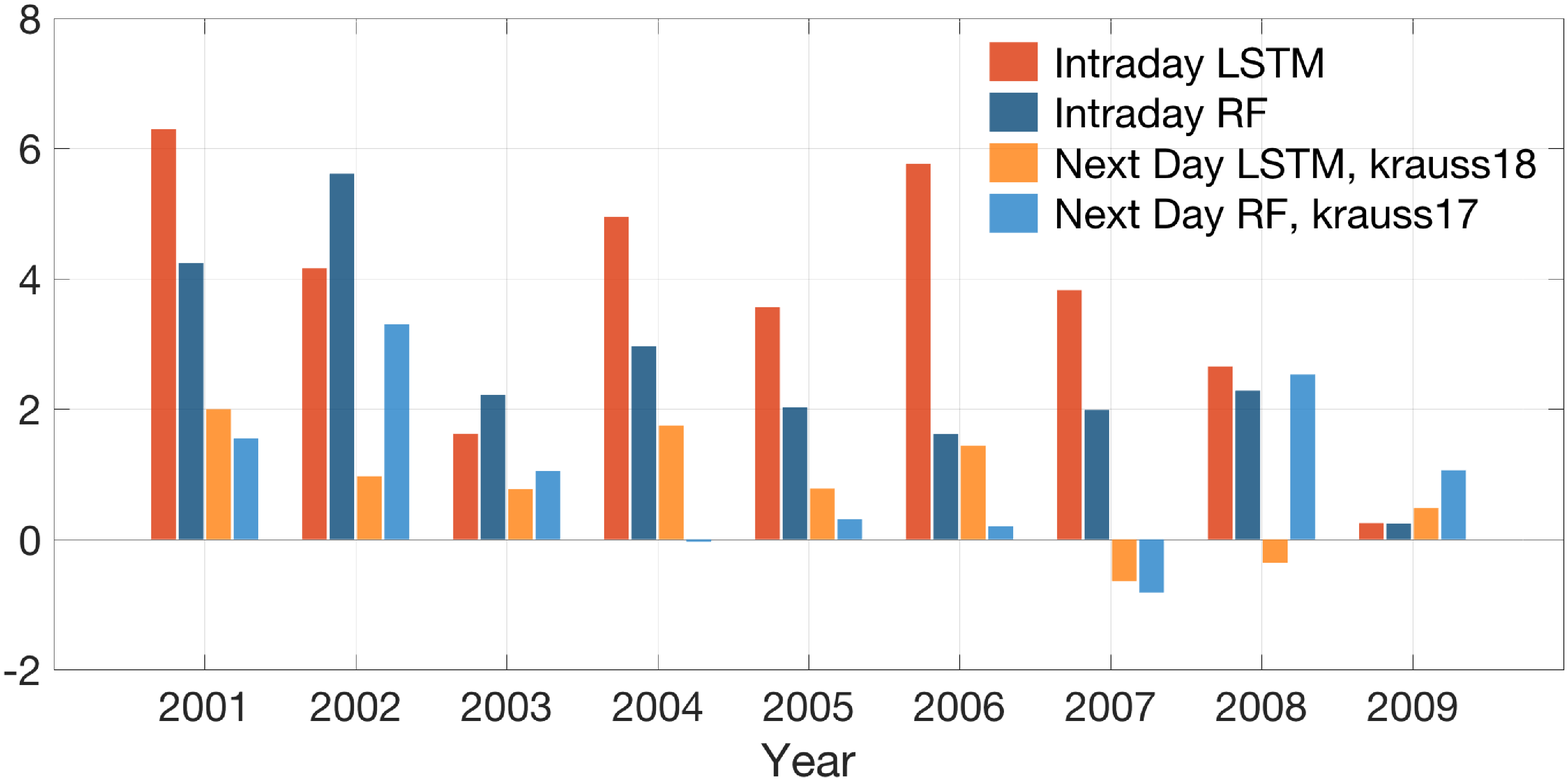} 
\hfill
\includegraphics[width=.4\textwidth]{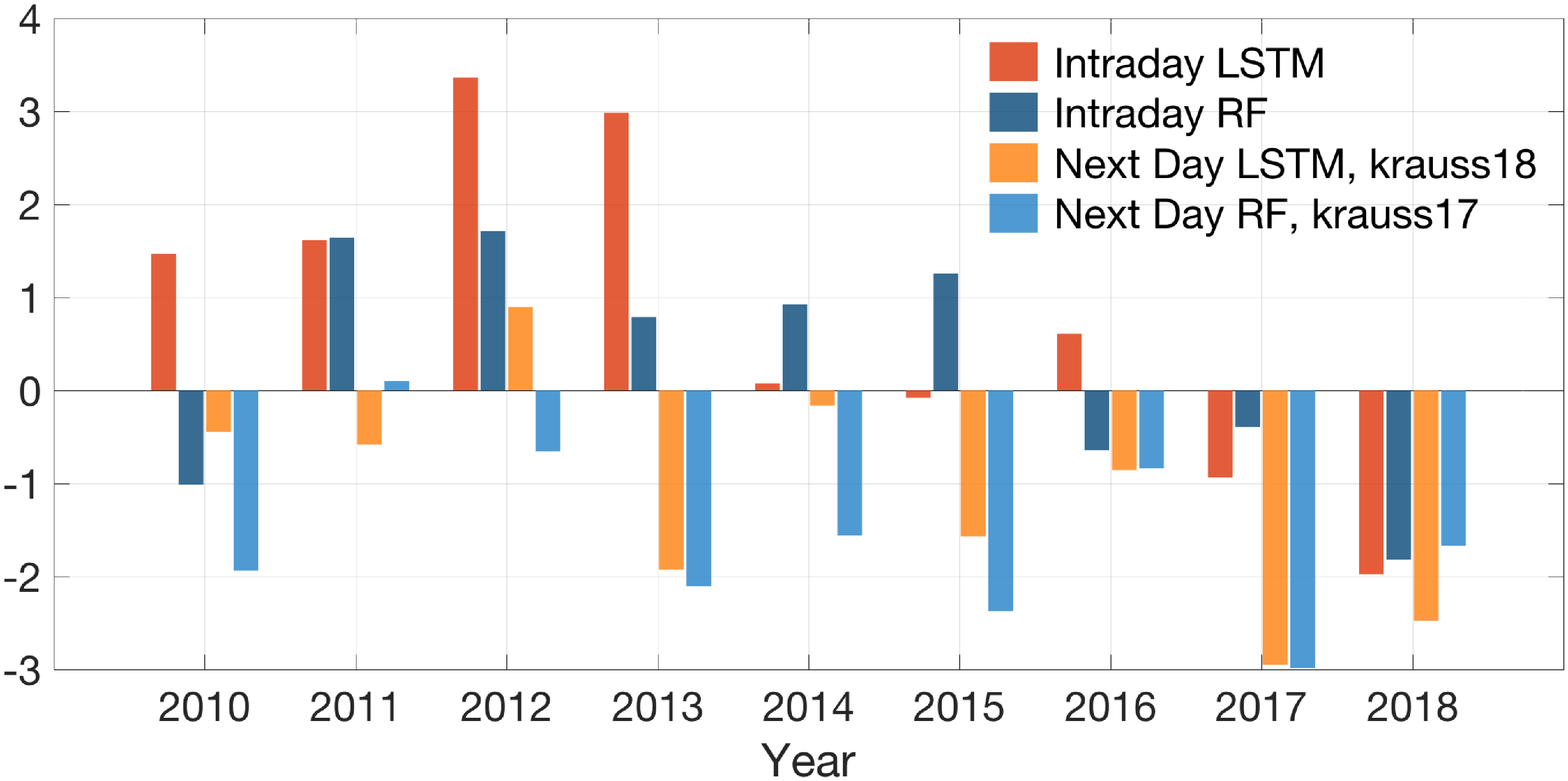} 
\caption{Annualised sharpe ratio, after deducting transaction cost}
\label{fig:figure3}
\end{figure}

\end{landscape}




\end{document}